%% file: ms.tex
\documentclass[letterpaper, 10 pt, conference]{ieeeconf}

\usepackage[hidelinks]{hyperref}

\input{tex/acronyms.tex}
\input{tex/math_aliases.tex}
\input{tex/helper_commands.tex}

\input{tex/my_ieee_stuff.tex}

\title{
\LARGE \bf
 \replaced[id=3, remark=5]{A Fast}{An Online} Trajectory Optimization Framework 
 for Legged Robots unifying ZMP and Capture Point Approaches
}

\title{
\LARGE \bf
 Fast Trajectory Optimization for Legged Robots \\
 using Vertex-based ZMP Constraints
}

\author{
Alexander W. Winkler, Farbod Farshidian, Diego Pardo, Michael Neunert, Jonas Buchli
\thanks{
Agile \& Dexterous Robotics Lab, Institute of Robotics
and Intelligent Systems, ETHZ Zurich, Switzerland. 
email: \{winklera, farbodf, depardo, neunertm, buchlij\}@ethz.ch
}
}

\begin{document}
\maketitle
\input{tex/content.tex}

\bibliographystyle{./bib/IEEEtran_no_url.bst} 
\bibliography{./bib/IEEEabrv.bib,./bib/library.bib,./bib/library_new.bib}


\end{document}

%% file: tex/acronyms.tex
\usepackage{glossaries}
\newacronym{hyq}{HyQ}{Hydraulically Actuated Quadruped}
\newacronym{com}{CoM}{Center of Mass}
\newacronym{zmp}{ZMP}{Zero-Moment-Point}
\newacronym{cp}{CP}{Capture Point}
\newacronym{cop}{CoP}{Center of Pressure}
\newacronym{oc}{OC}{Optimal Control}           
\newacronym{to}{TO}{Trajectory Optimization}
\newacronym{nlp}{NLP}{Nonlinear Programming Problem}
\newacronym{qp}{QP}{Quadratic Program}
\newacronym{cmaes}{CMA-ES}{Covariance Matrix Adaptation Evolution Strategy}
\newacronym{dof}{DoF}{Degrees of Freedom}
\newacronym{mpc}{MPC}{Model Predictive Control}
\newacronym{eom}{EoM}{Equation of Motion}
\newacronym{ip}{LIP}{Linear Inverted Pendulum}
\newacronym{imu}{IMU}{Inertial Measurement Unit}
\newacronym{ode}{ODE}{Ordinar Differential Equation}

%% file: tex/math_aliases.tex
\usepackage{amsmath}          
\usepackage{amssymb}          
\usepackage{siunitx}          
\usepackage{mathtools}        
\usepackage{xspace}
\usepackage{accents}          

\newcommand{\vect}[1]{\mathbf{#1}}
\newcommand{\dfdx}[2]{\frac{\partial{#1}}{\partial{#2}}}
\newcommand*\diff{\mathop{}\!\mathrm{d}} 
\newcommand{\dt}{\diff t}
\newcommand{\norm}[1]{\left\lVert #1 \right\rVert}
\newcommand{\real}[1]{\mathbb{R}^{#1}}
\newcommand{\integer}{\mathbb{Z}}
\newcommand{\ubar}[1]{\underaccent{\bar}{#1}}
\DeclareMathOperator*{\argmin}{arg\,min}
\newcommand{\pair}[2]{(#1,#2)}
\newcommand{\myin}{\,{\in}\,} 

\DeclarePairedDelimiterX\Set[1]{\lbrace}{\rbrace}%
 {  #1 }

\newcommand{\state}{\vect{x}}
\newcommand{\stated}{\dot{\vect{x}}}
\newcommand{\control}{\vect{u}}
\newcommand{\cost}{J}
\newcommand{\sic}{\vect{h}}            

\newcommand{\idxnode}{k\xspace}      

\newcommand{\idxlambda}{i}
\newcommand{\numlambdas}{n_u}

\newcommand{\idxfoot}{f}                 
\newcommand{\numfeet}{{n_\idxfoot}}
\newcommand{\contactflag}{c}              
\newcommand{\pxy}{\vect{p}^\idxfoot}       
\newcommand{\convexhull}{\mathcal{P}}     
\newcommand{\px}{p_x}                     
\newcommand{\eedd}{\ddot{\vect{p}}}       

\newcommand{\numcorners}{{n_v}}
\newcommand{\idxcorner}{v}             
\newcommand{\pxyv}{\vect{v}_\idxcorner}   

\newcommand{\tlocal}{\ubar{t}}       
\newcommand{\idxorder}{i}            
\newcommand{\numorder}{{n_\idxorder}} 
\newcommand{\powxy}{}                
\newcommand{\cxy}{\vect{c}\powxy}
\newcommand{\cx}{c_x}
\newcommand{\cxd}{\dot{c}_x}
\newcommand{\cxyd}{\dot{\vect{c}}\powxy}
\newcommand{\cxydd}{\ddot{\vect{c}}\powxy}

\newcommand{\uxy}{\vect{u}\powxy}       
\newcommand{\ux}{u_x}                     

\newcommand{\numpoly}{n}                  
\newcommand{\idxpoly}{k}                  

\newcommand{\decision}{\vect{w}}          
\newcommand{\comdecision}{\decision_c}    
\newcommand{\feetdecision}{\decision_p}    
\newcommand{\inputdecision}{\decision_u}    
\newcommand{\idxstep}{s\xspace}           
\newcommand{\numstepsperleg}{{n_\idxstep}}  
\newcommand{\bl}{\boldsymbol{\lambda}}    
\newcommand{\rangeset}{\mathcal{R}}       
\newcommand{\support}{support area~}       
\newcommand{\poly}{\vect{a}}              
\newcommand{\func}{\vect{f}}              
\newcommand{\lambdacost}{J_\lambda}

\newcommand{\qvec}{\vect{q}}                  
\newcommand{\qvecdd}{\ddot{\qvec}}            
\newcommand{\base}{\qvec_b}                   
\newcommand{\basedd}{\ddot{\qvec}}            
\newcommand{\baseref}{\qvec_{b,ref}}          
\newcommand{\based}{\dot{\qvec}_b}            
\newcommand{\basedref}{\dot{\qvec}_{b,ref}}   
\newcommand{\baseddref}{\ddot{\qvec}_{b,ref}} 
\newcommand{\jointddref}{\ddot{\qvec}_{j,ref}}
\newcommand{\jac}{\vect{J}}                   
\newcommand{\proj}{\vect{P}}                  
\newcommand{\rot}{\vect{R}}

%% file: tex/helper_commands.tex
\let\labelindent\relax \usepackage[inline]{enumitem} 
\newlist{mylist}{enumerate*}{1}
\setlist[mylist]{label=(\roman*)}

\usepackage{subfigure}

\usepackage{changes} 
\setauthormarkup{} 
\setremarkmarkup{[#1.#2]}
\definechangesauthor[name={General}, color={blue}]{gen}
\definechangesauthor[name={Editor}, color={blue}]{1}
\definechangesauthor[name={Reviewer 2}, color={blue}]{2}
\definechangesauthor[name={Reviewer 3}, color={orange}]{3}
\definechangesauthor[name={Reviewer 4}, color={blue}]{4}
\definechangesauthor[name={Reviewer 5}, color={blue}]{5}

\usepackage{hyperref}   

\usepackage{booktabs}

\newcommand{\fref}[1]{Fig.~\ref{#1}}
\newcommand{\eref}[1]{(\ref{#1})}
\newcommand{\tref}[1]{Table~\ref{#1}}
\newcommand{\aref}[1]{Appendix~\ref{#1}}

\usepackage{color}

%% file: tex/my_ieee_stuff.tex
\IEEEoverridecommandlockouts  
\newcommand{\citet}{\cite} 
\usepackage[noadjust]{cite} 

%% file: tex/content.tex
\begin{abstract}
This paper combines the fast \gls{zmp} approaches that work 
well in practice with the broader range of capabilities of a \acrlong{to} formulation,
by optimizing over body motion, footholds and \acrlong{cop} simultaneously.
We introduce a vertex-based representation of the support-area constraint,
which can treat arbitrarily oriented point-, line-, and area-contacts uniformly.
This generalization allows us to create motions such quadrupedal
walking, trotting, bounding, pacing, combinations and transitions between these, limping, 
bipedal walking and push-recovery all with the same approach. 
This formulation constitutes a minimal representation of the physical laws (unilateral contact forces)
and kinematic restrictions (range of motion) in legged locomotion, which      
allows us to generate various motion in less than a second.
We demonstrate the feasibility of the generated motions on a real quadruped robot.
\end{abstract}
\glsresetall 

\IEEEpeerreviewmaketitle

\section{Introduction}
\label{sec:intro}
Planning and executing motions for legged systems is a complex task. 
A central difficulty is that legs cannot pull on the ground, e.g. the forces acting on
the feet can only push upwards. Since the motion of the body is mostly
generated by these constrained (=unilateral) contact forces, this motion
is also restricted. When leaning forward past the tip of your tows, you 
will fall, since your feet cannot pull down to generate a momentum
that counteracts the gravity acting on your \gls{com}. Finding motions that
respect these physical laws can be tackled by various approaches described in the following.

A successful approach to tackle this problem is through full-body \gls{to}, in which an optimal
body and endeffector motion plus the appropriate inputs are discovered to achieve a high-level goal.
This was demonstrated by \cite{Schultz2010,Dai2014,Posa2013,mordatch2012,Posa2016a,Neunert2017, Pardo2017} resulting
in an impressive range of motions for legged systems.
These \gls{to} approaches have shown great performance, but are often time consuming to calculate and
not straight-forward to apply on a real robot. 
In \cite{Coros2011} the authors generate an wide range of quadruped gaits, transitions and
jumps based on a parameterized controller and periodic motions. While the 
resulting motions are similar to ours, the methods are very different:
While our approach is based on \gls{to} with physical constraints, \cite{Coros2011}
optimizes controller parameters based mainly on motion capture data.

Previous research has shown that to generate feasible motions to
execute on legged systems, non-\gls{to} approaches also work well, although the motions 
cannot cover the range of the approaches above. One way is to model the robot
as a \gls{ip} and 
keep 
the \gls{zmp} \cite{2004_vukobratovic} inside the convex hull of the feet in stance.
This approach has been successfully
applied to generate motions for biped and quadruped walking \cite{Kajita2003b,Kalakrishnan2010a,Kolter2008,Carpentier2016}.
However, these hierarchical approaches use predefined footholds, usually
provided by a higher-level planner beforehand that takes terrain information (height, slope) into account.
Although this decoupling of foothold planning and body motion generation reduces complexity,
it is unnatural, as the main intention of the footholds is to assist the body to achieve a desired motion. 
By providing fixed foot-trajectories that the body motion planner cannot modify, 
constraints such as stability or kinematic reachability become purely the responsibility of the lower-level body motion planner,
artificially constraining the solution.
%
A somewhat reverse view of the above are \acrfull{cp} \cite{pratt2006}
approaches, which have been successfully used to generate dynamic
trotting and push recovery motions for quadruped robots \cite{Gehring2013,
Barasuol2013a}. A desired body motion (usually a reference \gls{com} velocity)
is given by a high-level planner or heuristic, and a foothold/\gls{cop}
trajectory must be found that generates it. 
 
Because of the dependency between footholds and body motion,
approaches that optimize over both these quantities simultaneously, while still using
a simplified dynamics model, have been developed \cite{Mordatch2010,Stephens2010,Diedam2009,Herdt2010a,Naveau2017,winkler17}. 
This reduces heuristics while increasing 
the range of achievable motions, but still keeps computation time short compared to
full body \gls{to} approaches.
These approaches are most closely related to the work presented in this paper.

The approaches \cite{Stephens2010,Diedam2009,Herdt2010a,Naveau2017}
demonstrate impressive performance on biped robots. One common difficulty in these
approaches however is the nonlinearity of the \gls{cop} constraint with respect to the 
orientation of the feet. In \cite{Diedam2009,Herdt2010a} the orientation is either fixed or 
solved with a separate optimizer beforehand. In \cite{Naveau2017} the nonlinearity of this
constraint is accepted and the resulting nonlinear optimization problem solved. However, although
the orientation of the individual feet can be optimized over in these approaches, a combined
support-area with multiple feet in contact is often avoided, by not sampling the constraint
during the double-support phase. 
For biped robots neglecting the constraint in the double-support phase is not so critical,
as these take up little time during normal walking. For quadruped robots however, there are
almost always two or more feet in contact at a given time, so the correct representation of 
the dynamic constraint in this phase is essential.

We therefore extend the capabilities of the approaches above by using a vertex-based representation
of the \gls{cop} constraint, instead of hyperplanes. In \cite{Serra2016} this idea is briefly 
touched, however the connection between the corners of the foot geometry and the convexity variables
is not made and thereby the restriction of not sampling in the double-support phase remains.
Through our proposed formulation, double- and single-stance
support areas can be represented for arbitrary foot geometry, including point-feet.
Additionally, it allows to represent arbitrarily oriented 1D-support lines, which wasn't possible with the above approaches. Although
not essential for biped walk on flat feet, this is a core necessity for dynamic quadruped motions (trot, pace, bound).
This is a reason why \gls{zmp}-based approaches were so far only used for quadrupedal walking, where
2D-support areas were present.

The approach presented in this paper combines the \gls{ip}-based \gls{zmp} approaches that are fast and work 
well in practice with the broader range of capabilities of a \gls{to} formulation.
A summary of the explicit contributions with respect to the papers above are:
\begin{itemize}
\item We reformulate the 
traditional \gls{zmp}-based legged locomotion problem \cite{Kajita2003b}
into a standard \gls{to} formulation with the \gls{cop} as input, clearly 
identifying state, dynamic model and path- and boundary-constraints, which 
permits easier comparison with existing methods in the \gls{to} domain.
Push recovery behavior also naturally emerges from this formulation.
\item We introduce a vertex-based representation of the \gls{cop} constraint, instead of hyperplanes.
which allows us to treat arbitrarily oriented point-, line-, and area-contacts uniformly. This enables us to generate
motions that are difficult for other \gls{zmp}-based approaches, such as 
bipedal walk with double-support phases, point-feet locomotion, various gaits as well
as arbitrary combinations and transitions between these.
\item Instead of the heuristic shrinking of support areas, we introduce a cost
term for uncertainties that improve the robustness of the planned motions.
\end{itemize}
We demonstrate that the problem can be solved for multiple steps in less than a second to
generate walking, trotting, bounding, pacing, combinations and transitions between these, limping, 
biped walking and push-recovery motions for a quadruped robot.
Additionally, we verify the physically feasibility of the optimized motions
through demonstration of walking and trotting on a real \SI{80}{\kg} hydraulic quadruped. 
%
\section{Method}
\label{sec:approach}
\subsection{Physical Model}
\begin{figure}[tb]
	\centering
	\includegraphics[width=0.99\columnwidth]{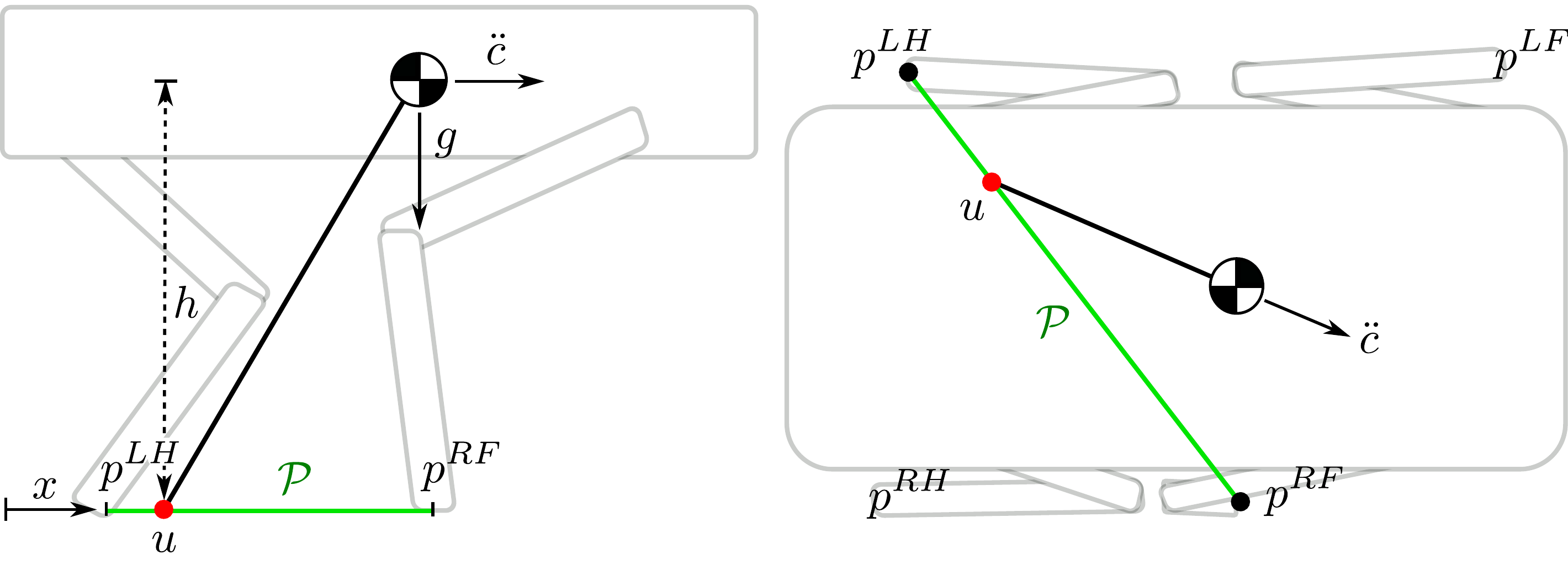}
	\caption{Modeling of a quadruped robot by a \gls{ip} with the right-front $\vect{p}^{RF}$ 
	 and left-hind $\vect{p}^{LH}$ legs in contact. Through joint torques the robot can control the 
	center of pressure $\uxy$ and thereby the motion of the
	\gls{com} $\cxydd$. However, $\uxy$ can only lie inside
	the convex hull (green line) of the contact points.}
	\label{fig:inverted_pendulum_top}
\end{figure}

We model the legged robot as a \acrfull{ip},
with its \gls{com} ${\cxy\!=\!\pair{\cx}{c_y}}$ located at a constant height $h$.
The touchdown position of the pendulum with the ground (also known as \gls{zmp} or \gls{cop})  
is given by ${\uxy\!=\!\pair{\ux}{u_y}}$ as seen in \fref{fig:inverted_pendulum_top}.
The \gls{com} acceleration $\cxydd$ is predefined by the physics of
a tipping pendulum
\begin{align}
\begin{bmatrix}
\cxyd \\
\cxydd
\end{bmatrix}
&=
\func(\state,\control)
=
\begin{bmatrix}
\cxyd \\
(\cxy - \uxy) gh^{-1}
\end{bmatrix}
\label{eq:lip_ode}.
\end{align} 
The second-order dynamics are influenced by the \gls{com} position $\cxy$, the \gls{cop} $\control$ and gravity $g$.
This model can be used to describe a legged robot, since the robot can 
control the torques in the joints, thereby the contact forces and through these the 
position of the \gls{cop}. Looking only at the x-direction (left image in \fref{fig:inverted_pendulum_top}), if the robot 
decides to lift the hind leg, the model 
describing the system dynamics is a pendulum in contact with the ground at
the front foot $\vect{p}^{RF}$, so ${\uxy \!=\! \vect{p}^{RF} \!=\! \pair{\px}{p_y}}$. Since this pendulum is nearly upright, the 
\gls{com} will barely accelerate in x; the robot is balancing on the front leg. However,
lifting the front leg can be modeled as placing the pendulum at ${\uxy = \vect{p}^{LH}}$, 
which is strongly leaning and thereby must accelerate forward in x. By distributing
the load between the legs, the robot can generate motions corresponding to a 
pendulum anchored anywhere between the contact points, e.g. $\uxy \in \convexhull$ (see \fref{fig:inverted_pendulum_top}).
Therefore, the \gls{cop} $\uxy$ is considered the input to the system 
and an abstraction of the joint torques and contact forces.

\subsection{Trajectory Optimization Problem}
We want to obtain the inputs $\control(t)$ that generate a motion $\state(t)$ from an
initial state $\state_0$ to a desired goal state $\state_T$ in time $T$ for a robot described by the system dynamics $\func(\state,\control)$,
while respecting some constraints $\sic(\state,\control) \leq 0$ and optimizing a performance criteria $\cost$.
This can be formulated as a continuous-time \gls{to} problem 
\begin{subequations}
\begin{align}
\text{find \hspace{1.3cm}} \state(t), \control(t)& , \quad \text{for } t \in [0,T] \\
\text{subject to} \hspace{0.5cm} \state(0) - \state_0                      &= 0       \quad \text{  (given initial state)}     \label{eq:initial_constraints} \\
                                  \stated(t) - \func(\state(t),\control(t))&= 0       \quad \text{  (dynamic model)}    \label{eq:model}\\          
                                  \sic(\state(t),\control(t))              &\leq 0    \quad \text{  (path constraints)} \label{eq:state_input_constraints}\\
                                  \state(T) - \state_T                     &= 0       \quad \text{  (desired final state)}    \label{eq:terminal_constraints}\\
                                  \state(t), \control(t) &= \argmin \cost(\state, \control)     \label{eq:cost_function}.                 
\end{align}
\label{eq:oc_formulation}
\end{subequations}
The dynamics are modeled as those of a \gls{ip} \eref{eq:lip_ode}, whereas
the state and input for the legged system model are given by
\begin{align}
\state(t) &=
\begin{bmatrix} 
\cxy & \cxyd &    \vect{p}^1,\alpha^1,\dots,\vect{p}^\numfeet,\alpha^\numfeet  
\end{bmatrix}^T \\
\control(t) &= \uxy
\label{eq:state_input_oc},
\end{align}
which includes the \gls{com} position and velocity and the position and orientation of the $\numfeet$ feet.
The input $\control(t)$ to move the system is the generated \gls{cop}, abstracting the usually used contact forces or joint torques.

\subsection{Specific Case: Capture Point}
We briefly show that this general \gls{to} formulation, using the \gls{ip} model,
also encompasses Capture Point methods to generate walking motions.
Consider the problem of finding the position to step with a point-foot robot
to recover from a push.
With the initial position $\cxy_0$ and the initial velocity $\cxyd_0$
generated by the force of the push we have ${\state_0\!=\!\pair{\cxy_0}{\cxyd_0}}$.
The robot should come to, and remain, at a stop at the end of the motion, irrespective of where and when,
so we have ${\cxyd_{T \to \infty}\!=\!0}$. We parametrize the input by the constant parameter 
${\control(t)\!=\!\uxy_0}$, as we only allow one step with a point-foot.
We allow the \gls{cop} to be placed anywhere, e.g. no path constraints \eref{eq:state_input_constraints}
and do not have a preference as to how the robot achieves this task, e.g. ${\cost(\state, \control)\!=\!0}$.

Such a simple \gls{to} problem can be solved analytically, 
without resorting to a mathematical optimization solver (see \aref{ap:capture_point}).
The point on the ground to generate and hold the \gls{cop} in order to achieve a final steady-state maintaining zero \gls{com} velocity 
becomes
\begin{equation}
\control(t) = \uxy_0 = \cxy_0 + \cxyd_0hg^{-1}. 
\end{equation}
This is the one-step \acrfull{cp}, originally derived by \cite{pratt2006} and 
the solution of our general \gls{to} formulation \eref{eq:oc_formulation} for a very 
specific case (e.g one step/control input, zero final velocity).

\subsection{General Case: Legged Locomotion Formulation}
\label{subsec:my_oc_highlevel}
\begin{figure}[tb]
	\centering
	\includegraphics[width=0.99\columnwidth]{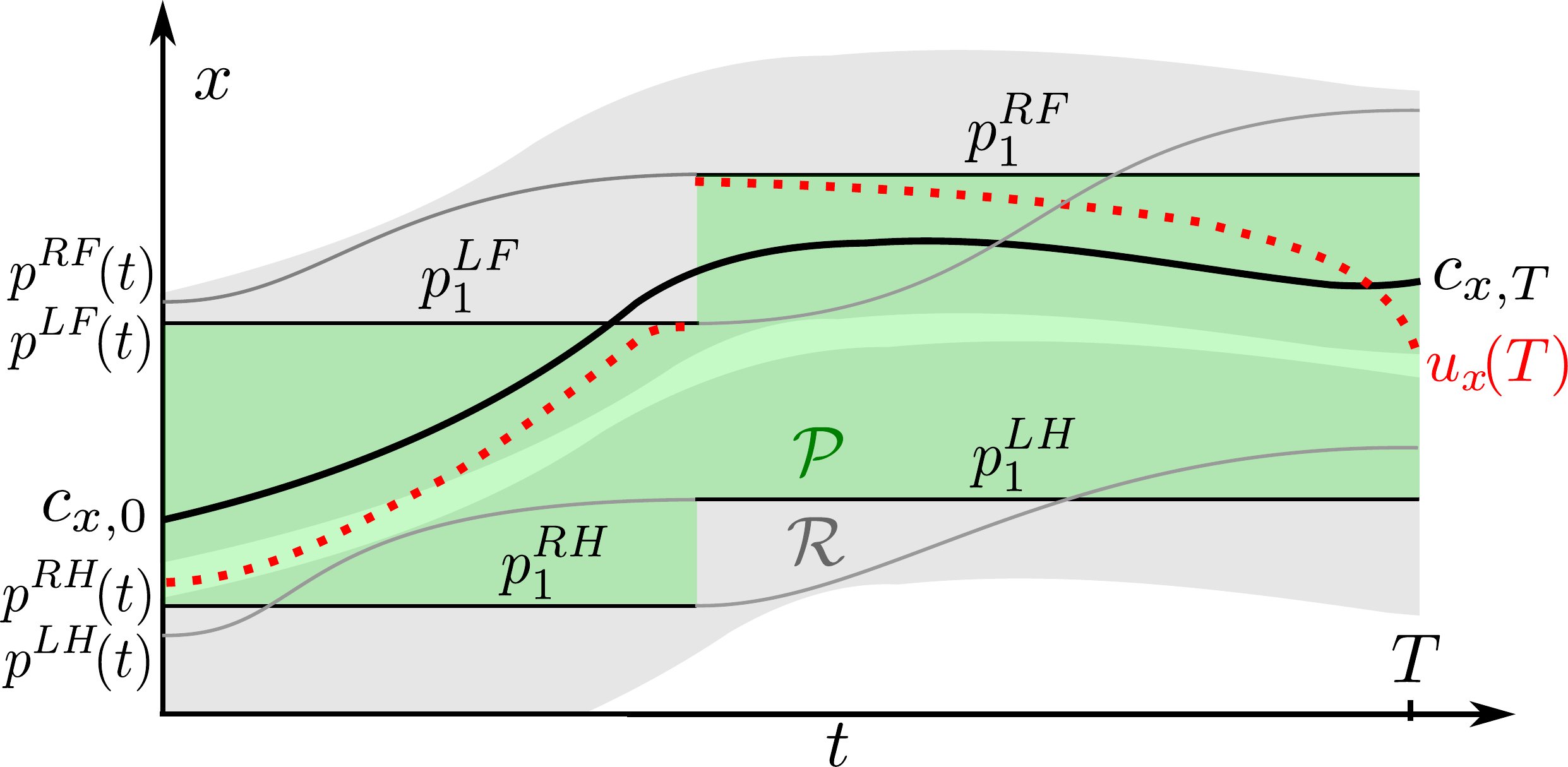}
	\caption{Overview of the \gls{to} problem: A point-foot quadruped robot 
	trotting forward in x-direction, 
	first swinging right-front and left-hind legs ${\idxfoot \in \{RF,LH\}}$, then left-front and right-hind ${\idxfoot \in \{LF,RH\}}$.
	The \gls{com} motion $\cx$ (black line) is generated by shifting the \gls{cop} $\ux(t)$ (red dots). However,
	$\ux$ can only lie in the convex hull $\convexhull$ (green area) of the legs in contact at that time $t$.
	Additionally, the position of each leg $p^\idxfoot$ must always be inside its range of 
	motion $\rangeset$ (gray areas for front and hind legs) relative to the \gls{com}. The optimization problem consist of varying 
	the position of the footholds $p^\idxfoot_\idxstep \in \rangeset$, 
	to allow inputs $\ux \in \convexhull$ that drive the robot from an initial position $c_{x,0}$ to a desired goal position $c_{x,T}$
	in time $T$. 
	}
	\label{fig:oc_state_control}
\end{figure}

Compared to the above example, our proposed formulation adds the capabilities to
represent motions of multiple steps, time-varying \gls{cop}, physical restrictions
as to where the \gls{cop} can be generated and preferences which of the feasible motions to choose. 
This \gls{to} formulation is explained on a high-level in the following, corresponding to \fref{fig:oc_state_control}, 
whereas more specific details of the implementation are postponed to the next section.

\subsubsection{Unilateral Forces}
We clearly differentiate between the \gls{cop} $\uxy$ and the feet positions $\pxy$, which only
coincide for a point-foot robot with one leg in contact. Generally, the footholds affect the input bounds
of $\control$. We use $\control$ to control the body, but must
at the same time choose appropriate footholds to respect the unilateral forces constraint. 
Traditional \gls{zmp} approaches fix the footholds $\pxy$ in advance, as the combination
of both the \gls{cop} $\uxy$ and the footholds make this constraint nonlinear. We accept this 
nonlinearity and the higher numerical complexity associated with it. 
This gives us a much larger range
of inputs $\uxy$, as we can ``customize'' our bounds $\convexhull$ by modifying the footholds according to the desired task.
Therefore, the first path constraint of our \gls{to} problem is given by
\begin{equation}
\sic_1(\state(t),\control(t)) \le 0
\quad \Leftrightarrow \quad
\control \in \convexhull(\pxy,\alpha^\idxfoot,\contactflag^\idxfoot)
\label{eq:oc_ip_cop},
\end{equation}
where $\convexhull$ represents the convex hull of the feet in contact as seen in \fref{fig:oc_state_control} and
$\contactflag^\idxfoot \in \{0,1\} \in \integer$ is the indicator if foot $\idxfoot$ is in contact.

We implement this convex hull constraint by weighing the vertices/corners of each foot in contact. This extends the
capabilities of traditional representations by line segments/hyperplanes to also model
point- and line-contacts of arbitrary orientation.
We use predefined contact sequences and timings $\contactflag^\idxfoot(t)$, to
only optimize over real-valued decision variables $w \in \real{}$ and not
turn the problem into a mixed-integer \gls{nlp}. Simply by adapting this contact schedule $\contactflag^\idxfoot(t)$,
the optimizer generates various gaits as well as combinations and transitions between these,
for which previously separate frameworks were necessary.



\subsubsection{Kinematic Reachability}
When modifying the footholds to enclose the \gls{cop}, we must additionally ensure that these stay inside
the kinematic range $\rangeset$ of the legs (Reachability). This constraint that depends on both the \gls{com} $\cxy$ and 
foothold positions $\pxy$  is formulated for every leg $\idxfoot$ as
\begin{equation}
\sic_2(\state(t)) \le 0
\quad \Leftrightarrow \quad
\pxy \in \rangeset(\cxy).
\label{eq:oc_ip_kinematic}
\end{equation}
Allowing the modification of both these quantities simultaneously characterizes the legged locomotion
problem more accurately and reduces heuristics used in hierarchical approaches.

\subsubsection{Robust Motions}
With the above constraints the motion will comply to physics and the kinematics of the system.
This feasible motion is assuming a simplified model, a perfect tracking controller
and an accurate initial state. 
To make solutions robust to real world discrepancies where these assumptions are violated, it is best
to avoid the borders of feasible solutions, where the inequality constraints are tight (${\sic\!=\!\vect{0}}$). 
This can be achieved by artificially shrinking the solutions space by a stability margin (e.g. ${\sic \! \le \! \vect{m}}$).
 For legged locomotion this is often done by
shrinking the support area to avoid solutions were the \gls{cop} is placed at the 
marginally-stable border \cite{Kalakrishnan2010a}. 


We do not restrict the solution space, but choose the more conservative of the feasible motions through
a performance criteria $\lambdacost$.
This soft constraint expresses ``avoid boundaries when possible, but permit if necessary''.
The robot is allowed to be at marginally stable states, but since there are many uncertainties in our model and assumptions,
it is safer to avoid them.
This cost does not require a hand-tuned stability margin and the solution can still be at the boundaries when necessary.
However, especially for slow motions (e.g. walking) where small inaccuracies
can accumulate and cause the robot to fall, this cost term is essential to generate robust motions
for real systems.

\section{Implementation}
\label{sec:opt_problem}
There exist different methods to solve Optimal Control problems \eref{eq:oc_formulation}, namely
Dynamic Programming (Bellman Optimality Equation), indirect (Maximum Principle) and 
direct methods \cite{Diehl2006}.
In direct methods the continuous time \gls{to} problem 
is represented by a finite number of decision variables
and constraints and solved by a nonlinear programming solver. 
If the decision variables $\decision$ fully describe the input $\control(t)$ and 
state $\state(t)$ over time, the method is further classified as a simultaneous
direct method, with flavors Direct Transcription and Multiple Shooting.
In our approach we chose a Direct Transcription formulation,
e.g. optimizing state and controls together. This has the advantage of not requiring an ODE solver, 
constraints on the state can be directly formulated and the sparse structure of the Jacobian often
improves convergence. The resulting discrete formulation 
to solve the continuous problem in \eref{eq:oc_formulation} is given by
\begin{align*}
\text{find \hspace{1.15cm}} &\decision = (\comdecision, \feetdecision, \inputdecision) \\
\text{subject to \hspace{0.2cm} }
&\eref{eq:initial_constraints},  \hspace{2.7cm} \text{(given initial state)}\\ 
&\eref{eq:com_continuity},
\eref{eq:com_accel_constraint},
\eref{eq:kinematic_constraints},
\eref{eq:convexity_all}          \hspace{0.5cm} \text{(path-constraints)}\\
&\eref{eq:terminal_constraints}, \hspace{2.7cm} \text{(desired final state)} \\
&\decision = \argmin \eref{eq:lambda_cost}, \hspace{0.8cm} \text{(robustness cost)} 
\end{align*}
where $\comdecision$ are the parameters describing the \gls{com} motion, 
$\feetdecision$ the feet motion (swing and stance) and
$\inputdecision$ the position of the \gls{cop}. 
This section describes in detail how we parametrize the state $(\comdecision,\feetdecision)$ and input $\inputdecision$,
formulate the path-constraints
and defined the cost \eref{eq:lambda_cost}.

\subsection{Center-of-Mass Motion}
This section explains how the continuous motion of the \gls{com} can be described by
a finite number of variables to optimize over, while ensuring compliance with the \gls{ip} dynamics.

\subsubsection{CoM Parametrization}
\renewcommand{\numorder}{4} 
\renewcommand{\tlocal}{(t\!-\!t_\idxpoly)}
The \gls{com} motion is described by a spline, strung together by
$\numpoly$ quartic-polynomials as 
\begin{align}
\state(t)
&= 
\begin{bmatrix}
\cxy(t) \\
\cxyd(t)
\end{bmatrix}
=
\sum_{\idxorder=1}^\numorder
\begin{bmatrix}
\tlocal \\
\idxorder
\end{bmatrix}
\poly_{\idxpoly,\idxorder} \tlocal^{\idxorder\!-\!1}
+
\begin{bmatrix}
\poly_{\idxpoly,0} \\
\vect{0}
\end{bmatrix}
\label{eq:polynomial}
\\
\comdecision &= 
\begin{bmatrix}
\poly_{1,0}, \dots, \poly_{1,\numorder},\dots, \poly_{\numpoly,0}, \dots, \poly_{\numpoly,\numorder}
\end{bmatrix},
%
\end{align}
with coefficients $\poly_{\idxpoly,\idxorder} \in \real{2}$
and $t_\idxpoly$ describing the global time at the start of polynomial $\idxpoly$.

We ensure continuity of the spline by imposing equal position and 
velocity at each of the $\numpoly\!-\!1$ junctions between polynomial $\idxpoly$ and $\idxpoly\!+\!1$, so
$\state[t_{k+1}^-] = \state[t_{k+1}^+]$. 
Using $T_\idxpoly = t_{\idxpoly+1}\!-\!t_\idxpoly$ we enforce
\begin{align}
\sum_{\idxorder=1}^\numorder
\begin{bmatrix}
T_\idxpoly \\
\idxorder
\end{bmatrix}
\poly_{\idxpoly,\idxorder} 
T_\idxpoly^{\idxorder-1}
+
\begin{bmatrix}
\poly_{\idxpoly,0} \\
\vect{0}
\end{bmatrix}
\!=\! 
\begin{bmatrix}
\poly_{\idxpoly+1,0} \\
\poly_{\idxpoly+1,1}
\end{bmatrix}.
\label{eq:com_continuity}
\end{align}

\subsubsection{Dynamic Constraint}
In order to ensure consistency between the parametrized motion and the dynamics of the system \eref{eq:lip_ode},
the integration of our approximate solution $\cxydd(t)$ must resemble
that of the actual system dynamics, so
\begin{align}
\int_{t_k}^{t_{k+1}} \cxydd(t) \dt
\approx
\int_{t_k}^{t_{k+1}} \func_2(\state(t), \control(t)) \dt 
.
\label{eq:int_constraint}
\end{align}
Simpson's rule states that if $\cxydd(t)$ is chosen as a $2^{nd}$-order polynomial (which is  
why $\cxy(t)$ is chosen as $4^{th}$-order) 
that matches the system dynamics $\func_2$
at the beginning, the center and at
the end, then \eref{eq:int_constraint} is bounded 
by an error proportional to $(t_{\idxpoly+1} - t_\idxpoly)^4$. Therefore we 
add the following constraints for each polynomial
\begin{align}
\cxydd[t] 
=
\func_2(\state[t], \control[t]), 
\quad
\forall t \in 
\begin{Bmatrix}
t_\idxpoly, \frac{t_{\idxpoly+1}-t_{\idxpoly}}{2}, t_{\idxpoly+1} 
\end{Bmatrix}
\label{eq:com_accel_constraint}
\end{align}
%
(see \aref{ap:dynamic_constraint} for a more detailed formulation). By keeping
the duration of each polynomial short ($\sim$\SI{50}{ms}), the error of 
Simpson's integration stays small and the $\numorder^{th}$-order
polynomial solution $\cxy(t)$ is close to an actual solution of the
\gls{ode} in \eref{eq:lip_ode}. 

This formulation is similar to the ''collocation'' constraint \cite{Hargraves1987}.
Collocation implicitly enforces the constraints \eref{eq:com_accel_constraint} at the 
boundaries through a specific parametrization of the polynomial,
while the above formulation achieves this 
through explicit constraints in the \gls{nlp}. Reversely, collocation
enforces that $\dfdx{\cxy(t)}{t} = \cxyd(t)$ through the explicit constraint, 
while our formulation does this through parametrization in
\eref{eq:polynomial}.

\subsection{Feet Motion}
\subsubsection{Feet Parametrization}
We impose a constant position $\pxy_\idxstep$ and orientation $\alpha^\idxfoot_\idxstep$
if leg $\idxfoot$ is in stance.
We use a cubic polynomial in the ground plane to move the feet between two consecutive contacts
\newcommand{\tswing}{T_\idxstep} 
\begin{align}
\begin{bmatrix}\pxy(t) \\ \alpha^\idxfoot(t) \end{bmatrix}
=
\sum_{\idxorder=0}^3
\begin{bmatrix}
\poly_{\idxstep,\idxorder}^{\idxfoot} \\
b_{\idxstep,\idxorder}^{\idxfoot}
\end{bmatrix}
(t\!-t_\idxstep)^\idxorder,
\label{eq:foot_poly}
\end{align}
where $(t\!-t_\idxstep)$ is the elapsed time since the beginning of the swing motion.
The vertical swingleg motion does not affect the \gls{nlp} and 
is therefore not modeled.
The coefficients $\poly_{\idxstep,\idxorder} \in \real{2}$ and $b_{\idxstep,\idxorder} \in \real{}$ are fully
determined by the predefined swing duration and the 
position and orientation
of the enclosing contacts
$\begin{Bmatrix}\pxy_\idxstep,\alpha^\idxfoot_\idxstep \end{Bmatrix}$
and
$\begin{Bmatrix}\pxy_{\idxstep+1},\alpha^\idxfoot_{\idxstep+1} \end{Bmatrix}$.
%
%
\newcommand{\pdec}[2]{\vect{p}^#1_#2, \alpha^#1_#2} 
Therefore the continuous motion of all $\numfeet$ feet can be parametrized by
the \gls{nlp} decision variables
\begin{align}
\begin{split}
&\feetdecision = 
\begin{bmatrix}
\feetdecision^1  ,\dots,  \feetdecision^\numfeet
\end{bmatrix},\\
\text{where }
&\feetdecision^\idxfoot =
\begin{bmatrix}
\pdec{\idxfoot}{1},\dots, \pdec{\idxfoot}{\numstepsperleg}
\end{bmatrix}
\end{split}
\end{align}
are the parameters to 
fully describe the motion of a single leg $\idxfoot$ taking 
$\numstepsperleg$ steps.

\subsubsection{Range-of-Motion Constraint}
\label{subsec:kinematics}
\begin{figure}[tb]
	\centering
	\includegraphics[width=0.99\columnwidth]{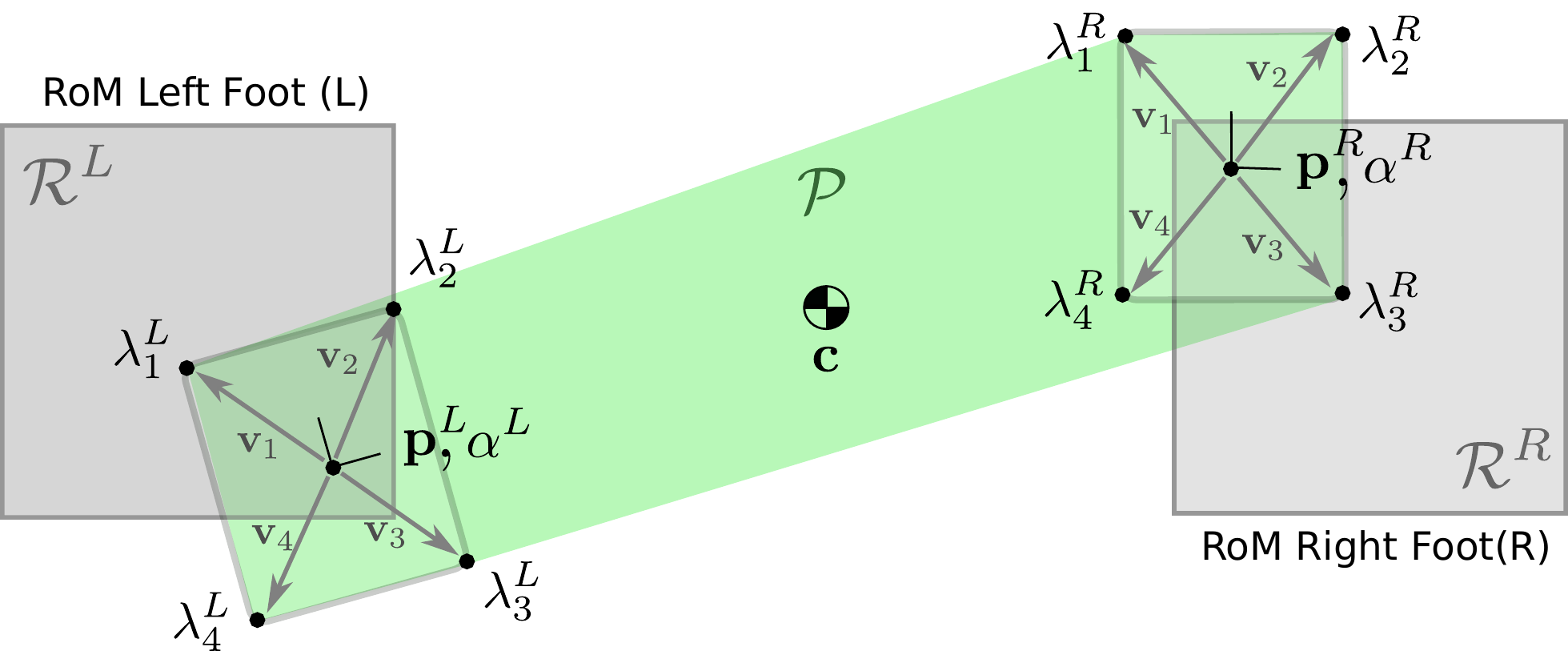}
	\caption{
	Top down view of a biped for both feet in contact at
	$\vect{p}^R,\vect{p}^L \in \rangeset$ inside the range of motion $\rangeset$ (gray), which 
	moves with the \gls{com} position.
	For square feet with corners $\pxyv$, rotated by $\alpha$, the support area is shown
	by $\convexhull$ (light green area). This is the area to which the \gls{cop} $\control$
	is constrained.
	If the biped controls its \gls{cop} to lie
	on the tip of the right foot, the corresponding corner
	carries all the load ($\lambda^R_1 = 1.0$), while the other seven lambdas are zero.
	In case of point-feet the \support is simply a straight line between $\vect{p}^R$ and $\vect{p}^L$.
	}
	\label{fig:rom_biped}
\end{figure}

\newcommand{\pnom}{\pxy_{nom}}
To ensure a feasible kinematic motion, we must enforce $\pxy \in \rangeset(\cxy)$, which is the gray area in \fref{fig:rom_biped}.
We approximate the area reachable by each foot through a rectangle $[-\vect{r}^{x,\!y}, \vect{r}^{x,\!y}]$, representing the allowed distance
that a foot can move from its nominal position $\pnom$ (center of gray area).
The foothold position for each foot $\idxfoot$ is therefore
constrained by
\begin{align}
  \label{eq:kinematic_constraints}
   -\vect{r}^{x,\!y} < \pxy[t] - \cxy[t] -  \pnom < \vect{r}^{x,\!y}.
\end{align}
Contrary to hierarchical approaches, this constraint allows the optimizer to either move the body to respect kinematic limits
\emph{or} place the feet at different positions. A constraint on the foot orientation can be formulated equivalently.

\subsection{Center of Pressure Motion}
To represent the continuous \gls{cop} trajectory, we parameterize it through the load carried by each
endeffector. This parametrization is used to formulate a novel convexity constraint based on vertices 
instead of hyperplanes. Finally this section introduces a cost that keeps the \gls{cop} from marginally
stable regions and improves robustness of the motion.

\subsubsection{CoP Parametrization}
The \gls{cop} $\uxy(t)$ is not parametrized by polynomial coefficients or discrete points, 
but by the relative load each corner of each foot is carrying. This load is given by
\begin{align}
\begin{split}
\bl(t) &= 
\begin{bmatrix}
\bl^1(t), \dots, \bl^\numfeet(t)
\end{bmatrix}^T, \\
\text{where }
\bl^\idxfoot(t) &= 
\begin{bmatrix}
\lambda_1^\idxfoot(t), \dots, \lambda^\idxfoot_\numcorners(t)
\end{bmatrix}
\in \left[0,1\right]^\numcorners.
\end{split}
\end{align}
$\numcorners$ represents the number of vertices/corners of foot $\idxfoot$.
For the square foot in \fref{fig:rom_biped}, four lambda values represent one foot and 
distribute the load amongst the corners.
These multipliers represent the percentage of vertical force that each foot
is carrying, e.g. $|\!|\bl^\idxfoot(t)|\!|_1=0.9$ implies that
leg $\idxfoot$ is
carrying 90\% of the weight of the robot at time $t$.
Using these values, the \gls{cop} is parameterized by 
\begin{align}
&\uxy(t) 
=\sum_{\idxfoot = 1}^\numfeet
\sum_{\idxcorner=1}^{\numcorners} 
\lambda_{\idxcorner}^\idxfoot(t)
(\pxy(t) + \rot(\alpha^\idxfoot(t)) \pxyv),
\label{eq:convexity1}
\end{align}
where $\rot(\alpha^\idxfoot) \in \mathbb{R}^{2 \times 2}$ represents the rotation matrix corresponding
to the optimized rotation $\alpha^\idxfoot$ of foot $\idxfoot$ \eref{eq:foot_poly}.
$\pxyv$ represents the fixed position (depending on the foot geometry) of
corner $\idxcorner$ of the foot expressed in the foot frame.
For a point-foot
robot with ${\pxyv\!= \!\vect{0}}$, \eref{eq:convexity1} simplifies to
$
\uxy = 
\sum_{\idxfoot = 1}^\numfeet \lambda^\idxfoot \pxy
$.

We represent $\bl(t)$ for the duration of the motion by piecewise-constant values $\bl_\idxlambda = \bl(t_\idxlambda)$ discretized every 
\SI{20}{ms}, resulting in $\numlambdas$ nodes. Therefore the \gls{cop} $\uxy$ can be fully parameterized by 
$\feetdecision$ and the additional
\gls{nlp} decision variables
\begin{align}
\inputdecision = 
\begin{bmatrix}
\bl_1, \dots, \bl_{\numlambdas}
\end{bmatrix}.
\label{eq:all_lambdas}
\end{align}

\subsubsection{Unilateral Forces Constraint}
We represent the essential input constraint \eref{eq:oc_ip_cop},
which ensures that only physically feasible forces inside the convex 
hull of the contacts are generated, for $\idxlambda = 1,\dots,\numlambdas$ as
\begin{subequations}
\begin{align}
\norm{\bl_\idxlambda}_1 &= 1,
\label{eq:convexity2} \\
0 \le \lambda_\idxcorner^\idxfoot[t_\idxlambda] &\le \contactflag^\idxfoot[t_\idxlambda],
\label{eq:convexity3}
\end{align}
\label{eq:convexity_all}
\end{subequations}
where $\contactflag^\idxfoot \myin \{0,1\} \myin \integer$ is the indicator if foot $\idxfoot$ is in contact.
The constraints \eref{eq:convexity1} and \eref{eq:convexity2} allow $\uxy$ to be located anywhere inside
the convex hull of the vertices of the current foot positions, independent of whether
they are in contact. However, since only feet in contact can actually carry load, \eref{eq:convexity3} enforces that
a leg that is swinging (${\contactflag^\idxfoot\!=\!0}$) must have all the corners of its
foot unloaded. These constraints together ensure that the \gls{cop} lies inside the green area shown in \fref{fig:rom_biped}.

\subsubsection{Robust Walking Cost}
\label{subsec:cost_lambda}
To keep the \gls{cop} away from the edges of the support-area we could constrain 
$\lambda_\idxcorner^\idxfoot$ of each leg in stance to be greater than a threshold,
 causing these legs in contact to never be unloaded.
This conceptually corresponds to previous approaches that
heuristically shrink support areas and thereby reduce the solution-space for \emph{all} situations. 
We propose a cost that has similar effect, but still permits the solver to use the limits of the 
space if necessary.

\newcommand{\nvt}{n_v}
The most robust state to be in,
is when the weight of the robot is equally distributed amongst
all the corners in contact, so
\begin{align}
\lambda^{\idxfoot*}_\idxcorner(t) = \frac{\contactflag^\idxfoot(t)}{\nvt(t)}, 
\end{align}
where $\nvt(t) = \numcorners \sum_{\idxfoot=1}^{\numfeet} \contactflag^\idxfoot(t)$ is the total number of vertices in contact at time $t$, predefined by the
contact sequence $\contactflag(t)$. This results in the \gls{cop} to be located in the center of the support
areas. 
The deviation of the input values from the optimal values $\lambda^*$ over the entire
discretized trajectory \eref{eq:all_lambdas} is then given by
\begin{align}
\lambdacost(\inputdecision) = 
  \sum_{\idxlambda=1}^{\numlambdas} 
  \norm{\bl_\idxlambda - \bl^*_\idxlambda}_2^2.
  \label{eq:lambda_cost}
\end{align}
For a support triangle (${\lambda^{\idxfoot*}_\idxcorner\!=\!\frac{1}{3}}$) this cost tries to keep
the \gls{cop} in the center and for a line (${\lambda^{\idxfoot*}_\idxcorner\!=\!\frac{1}{2}}$) in the middle.
For quadruped walking motions this formulation generates a smooth transition of the \gls{cop} between diagonally opposite swing-legs, 
while still staying away from the edges of support-areas whenever possible. 
%


\section{Tracking the Motion}
\begin{figure}[tb]
	\centering
	\includegraphics[width=0.99\columnwidth]{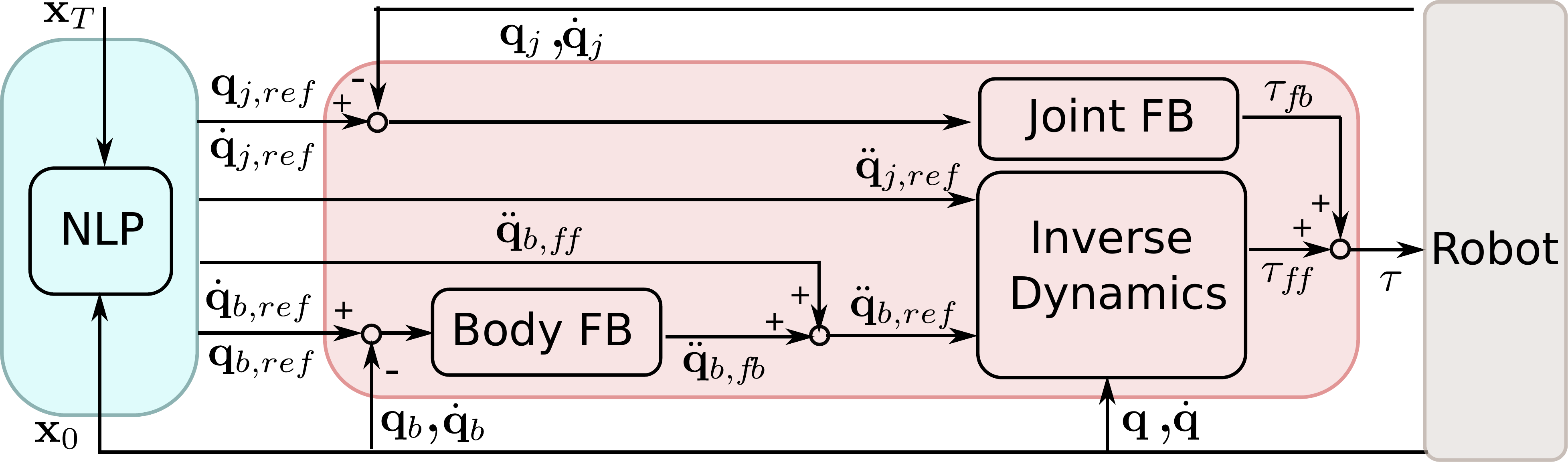}
	\caption{The controller that generates the required torques to execute a planned motion. Given
	the current state of the system $\state_0$ and a user defined goal state $\state_T$, the 
	optimizer generates a reference motion. We augment this reference through a body feedback
	acceleration based on how much the body deviates from the desired motion. Inverse dynamics
	is used to generate the torques to achieve the reference base and joint accelerations.}
	\label{fig:controller_overview}
\end{figure}

The motion optimization part of our approach is largely robot independent.
The only robot specific information needed to run the framework are the robot height,
the number of feet, their geometry and their kinematic range.
For execution however, the optimized motion must be translated into joint torques $\boldsymbol{\tau}$
using a fully-body dynamics model.
This section discusses this generation summarized by \fref{fig:controller_overview}.

\subsection{Generating full-body reference accelerations}
The 6--\gls{dof} base pose is reconstructed using zero desired orientation (in Euler angles x,y,z), the optimized \gls{com} motion $\cxy$
(assuming the geometric center of the base coincides with the \gls{com}),
and the constant base height $h$ as
\begin{equation*}
\baseref(t) = \begin{bmatrix}  0 & 0 & 0 & c_x(t) & c_y(t) & h \end{bmatrix}^T. 
\end{equation*}
In order to cope with uncertainties it is essential to
incorporate feedback into the control loop. We do this by adding an operational 
space PD-controller on the base that creates desired 6D base accelerations according to
\begin{equation*}
\baseddref = \basedd_{b,f\!f} + K_p(\base - \baseref) + K_d(\based - \basedref).
\end{equation*}
The derivate of the pose, the base twist $\based \in \mathbb{R}^6$ represents the
base angular and linear velocities and $\basedd_{b,f\!f}$ is the optimized \gls{com} 
acceleration from the \gls{nlp}. This controller modifies the planned body motion if the 
current state deviation from the reference state.

In order to obtain the desired joint accelerations that correspond to the planned
Cartesian motion of the feet we can use the relationship
$
\eedd(t) = \jac \ddot{\qvec} + \dot{\jac} \dot{\qvec},
$
where $\dot{\qvec},\ddot{\qvec} \in \real{6+n}$ represent the full body state (base + joints) and 
$
\jac = \begin{bmatrix} \jac_b & \jac_j \end{bmatrix} \in \mathbb{R}^{3\numfeet \times (6+n)}
$
the Jacobian that maps full-body velocities to linear foot velocities in world frame. 
Rearranging this equation, and using the Moore--Penrose pseudoinverse $\jac_j^+$, gives us
the reference joint acceleration
\begin{align}
 \jointddref = \jac_j^+ \left( \eedd -\dot{\jac} \dot{\qvec} - \jac_b \baseddref \right).
\label{eq:constraint_consistent_acc}
\end{align}
%

\subsection{Inverse Dynamics}
The inverse dynamics controller is responsible for generating required
joint torques $\boldsymbol{\tau}$ to track the reference acceleration $\qvecdd_{ref}$,
which is physically feasible based on the \gls{ip} model.   
This is done based on the rigid body dynamics model of the system, which 
depends on the joint torques, but also the unknown contact forces.
To eliminate the contact forces from the equation, we project 
it into the space of joint torques by
$
\proj = \vect{I} - \jac_c^+ \jac_c,
$
where $\vect{J}^T_c$ is the contact Jacobian that maps Cartesian
contact forces to joint torques \cite{Aghili2005a, Mistry2010}.
This allows us to solve for the required joint torques through
\begin{equation}
\boldsymbol{\tau} = (\proj \vect{S}^T)^+ \ \proj (\vect{M} \qvecdd_{ref} + \vect{C}),
\end{equation}
where  $\vect{M}$ is the joint space inertia matrix, $\vect{C}$ the effect of Coriolis forces on the
joint torques and $\vect{S}$ the selection matrix which prohibits from actuating the
floating base state directly.
We found it beneficial to also add a low-gain PD-controller on the 
joint position and velocities. This can mitigate the effects of dynamic modeling errors and
force tracking imperfections.

\section{Results}
\label{sec:results}
We demonstrate the performance of this approach on the hydraulically actuated
quadruped robot HyQ \cite{Semini2010}. The robot weighs approximately \SI{80}{\kg}, moves at a height of about \SI{0.6}{\m} and is 
torque controlled. Base estimation \cite{bloesch2013} is performed on-board, fusing \gls{imu} and joint
encoder values. Torque tracking is performed at \SI{1000}{\Hz}, 
while the reference position, velocity and torque set-points are provided at \SI{250}{\Hz}. 
The C++ dynamics model is generated by \cite{Frigerio2016}.
\newcommand{\imageheight}{2.5cm}
\newcommand{\imagewidth}{0.99\columnwidth}
\newcommand{\rotateAndInclude}[1]{\rotatebox{180}{\includegraphics{#1}}}
\newcommand{\motionfolder}{figs/rviz/motions_red}
\begin{figure}[tb]
	\centering
	\subfigure[Quadruped Walk: swinging one leg at a time.]{ %
	  \resizebox*{\imagewidth}{\imageheight}{ 
	    \rotateAndInclude{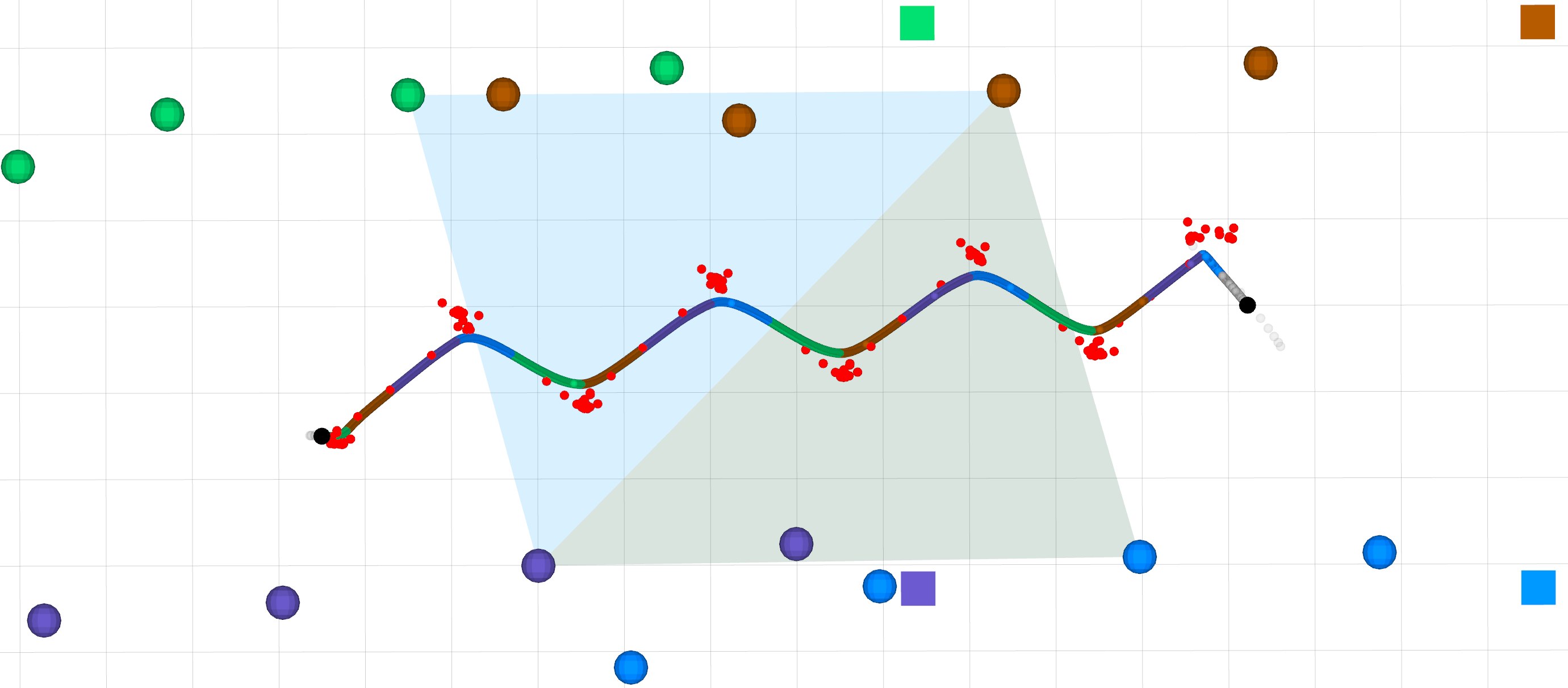}
	    \label{fig:com_motion_walk1}
	  }
	}
	\subfigure[Quadruped Trot: swinging diagonally opposite legs.]{ %
	  \resizebox*{\imagewidth}{\imageheight}{ 
	    \rotateAndInclude{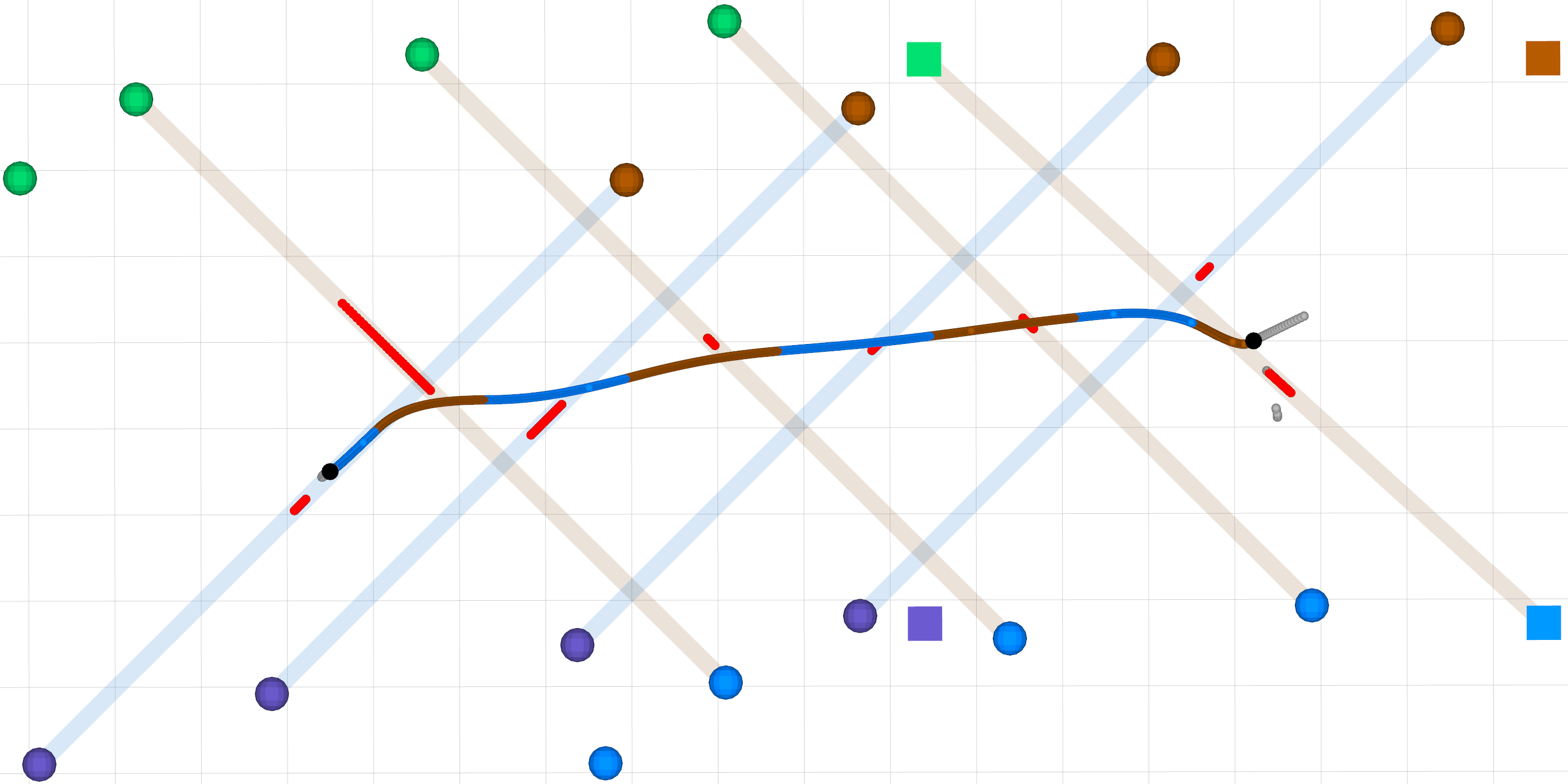}
	    \label{fig:com_motion_trot}
	  }
	}
	\subfigure[Quadruped Pace: swinging left, then right legs.]{ %
	  \resizebox*{\imagewidth}{\imageheight}{ 
	    \rotateAndInclude{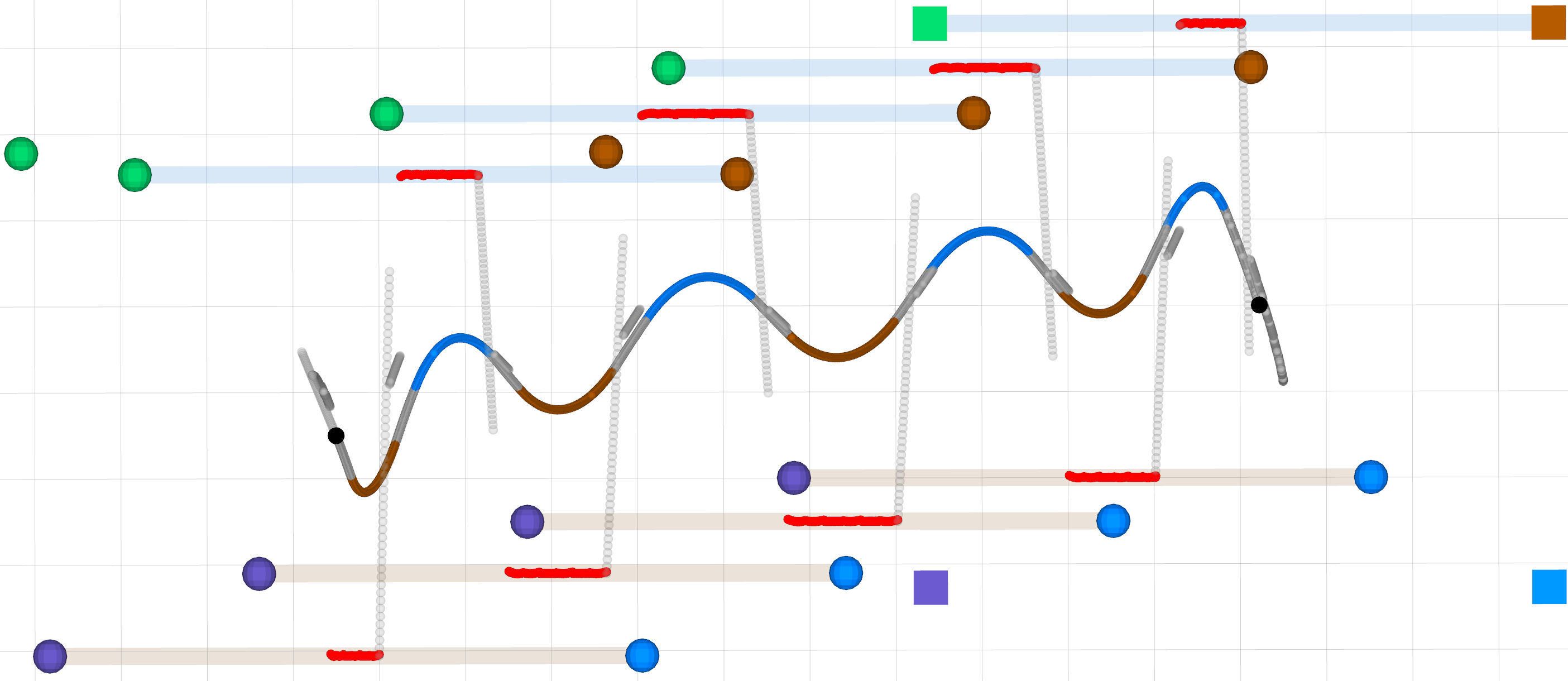}
	    \label{fig:com_motion_pace}
	  }
	}
	\subfigure[Quadruped Bound: swinging front, then hind legs.]{ %
	  \resizebox*{\imagewidth}{\imageheight}{ 
	    \rotateAndInclude{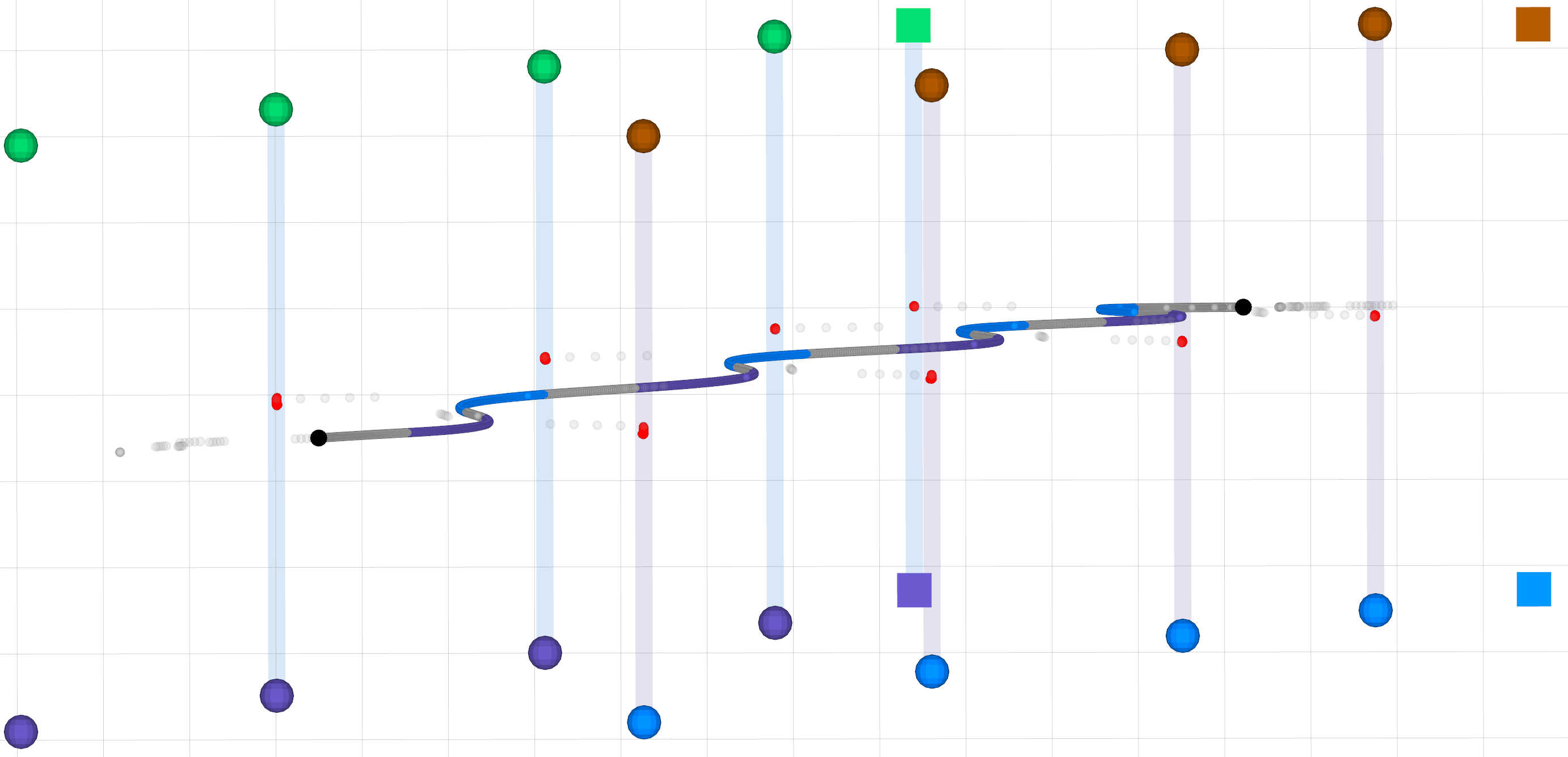}
	    \label{fig:com_motion_bound}
	  }
	}
	\caption{Top down view of the generated motions for a quadruped robot moving from left to right,
	swinging the legs $\idxfoot$ left-hind (blue), left-front (purple), right-hind (brown), right-front (green) in the 
	sequence shown. The initial stance is shown by the squares, the optimized steps by the circles.
	The \gls{com} motion $\pxy(t)$ is shown by the solid line, where the color corresponds 
	to the swingleg(s) at that moment. If all legs are in contact during, the motion and corresponding \gls{cop} is
	shown in gray. The \support for each phase is shown by the transparent areas. 
	The optimized \gls{cop} positions $\uxy(t)$ that drive the system are shown in red and always lie inside the support area.
	} 
	\label{fig:cop_motions}
\end{figure}

\subsection{Discussion of generated motions}
This section analyses the different motions generated by
changing the sequence and timings of contacts $\contactflag(t)$. There is no
high-level footstep planner; the footholds are chosen by the optimizer to 
enable the body to reach a user defined goal state $\state_T$. The results where obtained using C++ code interfaced 
with Interior Point Method (Ipopt \cite{Waechter2006})
or Sequential Quadratic Programming (Snopt \cite{Gill2002}) solvers
on an Intel Core i7/\SI{2.8}{GHz} Quadcore laptop. 
The Jacobians of the constraint and the gradient of the cost function are provided to the solver analytically, which
is important for performance. 
We initialize the decision variables $\decision$ with the quadruped standing in default stance for a given duration. 
The shown motions correspond to the first columns (e.g. 16 steps) in \tref{tab:results}.
The reader is encouraged to view the video at \url{https://youtu.be/5WLeQMBuv30}, as it very intuitively demonstrates the performance 
of this approach. Apart from the basic gaits, the video shows the capability of the framework to
generate gradual transitions between them, bipedal walking, limping and push-recovery. 

\subsubsection{Walk}
\fref{fig:com_motion_walk1} shows a walk of multiple steps, with the two support areas
highlighted for swinging RF$\to$LH. The effect of the cost term $\lambdacost$ is clearly visible, as
the \gls{cop} is accumulated away from the support area borders by left-right swaying of the body.
Only when switching diagonally opposite legs the \gls{cop} lies briefly at the marginally stable border, but then
immediately shifts to a more conservative location. Without the cost term, the \gls{com} motion
is a straight line between $\state_0$ and $\state_T$, causing the real system to fail. 
\subsubsection{Trot}
\fref{fig:com_motion_trot} shows a completely different pattern of support areas and \gls{cop}
distribution. During trotting only line-contacts exist, so the possible places to generate the 
\gls{cop} is extremely restricted compared to walking. Notice how the \gls{cop} lies 
close to the \gls{com} trajectory during the middle of the motion, but deviates quite large back/forward
during the start/end of the motion (e.g. the robot pushing off from the right-front (green) leg 
in the second to last step). This is because the distance between the \gls{cop} and
the \gls{com} generates the acceleration necessary for starting and stopping, whereas in the middle
the robot is moving with nearly constant velocity.  

\subsubsection{Pace/Bound/Biped Walk}
Specifying legs on the same side to be in contact, with a short four-leg transition period 
between them produces the motion shown in \fref{fig:com_motion_pace}. 
This can also be viewed as biped walking with line-feet (e.g. skis), with the constraint enforced
also \emph{during} the double-stance phase.
The first observation is the sideways swaying motion of the \gls{com}. This
is necessary because the support areas do not intersect (as in the trot) the \gls{com} trajectory.
Since the \gls{cop} always lies inside these left and right support areas,
they will accelerate the body away from that side until the next step, which then reverses the motion. 
We found that the \gls{ip} model with fixed zero body orientation
does not describe such a motion very well, as the inherent rotation (rolling) of the body is not taken into account.
In order to also demonstrate these motions on hardware,
the \gls{ip} model must be extended by the angular body motion.
Specifying the front and hind legs to alternate between contact generates a bound \fref{fig:com_motion_bound}.
The lateral shifting motion of the pace is now transformed to a forward backward motion of the \gls{com} due
to support areas. In case of an omni directional robot a bounding gait can simply be considered a side-ways pace.

\begin {table}[!t]
\caption{Specs of the \gls{nlp} for 16- and 4-step motions}
\label{tab:results} 
\centering
\begin{tabular}{@{} lcccc @{}}    
\toprule
&\multicolumn{4}{c}{(16 steps, \SI{1}{\m}) $|$ (4 steps, \SI{0.2}{\m})}  \\
\cmidrule{2-5}
\emph{} 	                                   & Walk                      & Trot              & Pace               &  Bound \\
\midrule
 Horizon $T$ [s]                    &6.4 $|$  1.6                   & 2.4 $|$ 0.6             & 3.2 $|$ 0.8           &  3.2 $|$ 0.8 \\
 Variables [-]                         & 646 $|$ 202                   & 387 $|$ 162                 & 1868 $|$ 728$\: \: $              & 1868 $|$ 728$\: \: $  \\
 Constraints [-]	  	                       & 850 $|$ 270                   & 548 $|$ 255                 & 2331 $|$ 939$\: \: $              & 2331 $|$ 939$\: \: $  \\
 $t_{\idxnode+1}-t_{\idxnode}$  [s]                    & 0.1                       & 0.05                & 0.02              & 0.02 \\
 Cost term                                   & $\lambdacost$               & -                   & -                 &  - \\
 \midrule
 Time Ipopt [s]	                       & 0.25 $|$ 0.06               & 0.02 $|$ 0.01           & 0.21 $|$ 0.12               &  0.17 $|$ 0.04  \\
 Time Snopt [s]  	                        & 0.35 $|$ 0.04            & 0.04 $|$ 0.01           & 0.54 $|$ 0.18               &  0.42 $|$ 0.29   \\
 \bottomrule
\end{tabular}
\end {table}
%

\section{Conclusion} 
\label{sec:conclusion}
This paper presented a \gls{to} formulation using vertex-based support-area
constraints, which enables the generation of a variety of motions for which previously
separate methods were necessary.
In the future, 
more decision variables (e.g. contact schedule, body orientation, foothold height
for uneven terrain),
constraints (e.g. friction cone, obstacles) and more sophisticated dynamic models can be incorporated into this formulation. 
Additionally, we plan to utilize the speed of the optimization for \gls{mpc}. 


\appendix
\subsection{Derivation of Capture Point}
\label{ap:capture_point}
Consider the differential equation describing a \gls{ip} (linear, constant coefficients, second order)
in x-direction
\begin{equation}
\ddot{c}(t) - \frac{g}{h}c(t) = -\frac{g}{h}u
\end{equation}
The general solution to the homogeneous part of the equation 
can be construct by the Ansatz
$
c(t) = e^{\alpha t} 
$
which leads to the characteristic equation
$
\alpha^2 e^{\alpha t} - \frac{g}{h} e^{\alpha t} = 0,
$
resulting in
$
\alpha = \pm \sqrt{\frac{g}{h}}.
$
Assuming constant input $u_0$ leads to the partial solution $c_p(t) = u_0$, 
and the space of solutions for the entire \gls{ode} is given by
\begin{equation}
c(t) = \beta_1 e^{\alpha t} + \beta_2 e^{-\alpha t} + u_0
\label{eq:ode_solution}
\end{equation}
where $\beta_1, \beta_2 \in \mathbb{R}$ are the free parameters describing the motion. Imposing
the initial position
$
c(0) = \beta_1 + \beta_2 + u_0         \overset{!}{=}  c_0 
$
and velocity
$
\dot{c}(0) = \alpha \beta_1 - \alpha \beta_2 \overset{!}{=}  \dot{c}_0 
$
we obtain
\begin{equation}
\beta_{1,2} = \frac{1}{2} (c_0 \pm \frac{\dot{c}_0}{\alpha} - u_0).
\label{eq:betas}
\end{equation}
As $t \to \infty$ we require the velocity to remain at zero (pendulum at rest).
Since $\lim_{t \to \infty} e^{-\alpha t} = 0$, and $\alpha \ne 0$ we have
\begin{align}
\lim_{t \to \infty} \cxd(t) &= \alpha \beta_1 \lim_{t \to \infty} e^{\alpha t} \overset{!}{=} 0
\quad \Leftrightarrow \quad \beta_1 = 0\\
&\Rightarrow u_0 = c_0 + \frac{h}{g}\dot{c}_0,
\end{align}
which is known as the one-step \acrlong{cp} originally derived in \cite{pratt2006}. 


\subsection{Dynamic Constraint}
\label{ap:dynamic_constraint}

\renewcommand{\tlocal}{\ubar{t}}
The system dynamics constraint \eref{eq:int_constraint} enforced through
$\cxydd[t] = \func_2(\state[t], \control[t])$, with the local polynomial time ${\tlocal\!=\!(t-t_\idxnode)}$,
are formulated as
%
\begin{align*}
%
&\cxydd[t] =
\sum_{\idxorder=2}^\numorder 
\idxorder (\idxorder-1) \poly_{\idxpoly,\idxorder} \tlocal^{\idxorder-2}
= \frac{g}{h}(\cxy(t) - \uxy(t)) \\
\Leftrightarrow
&\sum_{\idxorder=2}^\numorder 
\poly_{\idxpoly,\idxorder} \tlocal^{\idxorder-2}  \left( \idxorder (\idxorder-1) - \frac{g}{h} \tlocal^2 \right)
= \frac{g}{h} (\poly_{\idxpoly,0} + \poly_{\idxpoly,1}\tlocal -\uxy(t)).
\end{align*}

